# Automatic Search for Photoacoustic Marker Using Automated Transrectal Ultrasound

**Zijian Wu[1†], Hamid Moradi[2†], Shuojue Yang[3], Hyunwoo Song[1,4], Emad M. Boctor[1,4*], and Septimiu E. Salcudean[2*]**

[1]*Johns Hopkins University, Laboratory for Computational Sensing and Robotics, Baltimore, MD, USA*
[2]*University of British Columbia, Dept. of Electrical and Computer Engineering, Vancouver, BC, CA*
[3]*University of Texas at Austin, Walker Department of Mechanical Engineering, Austin, TX, USA*
[4]*Johns Hopkins University, Dept. of Computer Science, Baltimore, MD, USA*
[†]*Equal Authorship Contribution*
[*]*tims@ece.ubc.ca; eboctor@jhu.edu*

**Abstract:** Real-time transrectal ultrasound (TRUS) image guidance during robot-assisted laparoscopic radical prostatectomy has the potential to enhance surgery outcomes. Whether conventional or photoacoustic TRUS is used, the robotic system and the TRUS must be registered to each other. Accurate registration can be performed using photoacoustic (PA markers). However, this requires a manual search by an assistant [19]. This paper introduces the first automatic search for PA markers using a transrectal ultrasound robot. This effectively reduces the challenges associated with the da Vinci-TRUS registration. This paper investigated the performance of three search algorithms in simulation and experiment: Weighted Average (WA), Golden Section Search (GSS), and Ternary Search (TS). For validation, a surgical prostate scenario was mimicked and various *ex vivo* tissues were tested. As a result, the WA algorithm can achieve $0.53^{o} \pm 0.30^{o}$ average error after 9 data acquisitions, while the TS and GSS algorithm can achieve $0.29° \pm 0.31°$ and $0.48^{o} \pm 0.32^{o}$ average errors after 28 data acquisitions.

## 1. Introduction

Prostate cancer (PCa) is one of the most common malignant tumors among men worldwide, with a high mortality rate [1]. According to [2], 11.6% of men will develop prostate cancer in their lifetime, with approximately a 20% death rate in the United States. Radical prostatectomy is a popular surgical approach to treat PCa by removing the entire prostate gland since 1905 [3,4]. In clinical practice, the traditional open radical prostatectomy (ORP) has almost been replaced by laparoscopic radical prostatectomy (RLP) [5]. As a minimally invasive surgical procedure for PCa, RLP significantly reduces blood loss, hospitalization duration, and postoperative complications [6]. However, the long learning curve associated with laparoscopic procedures limits the application of RLP [7]. Robot-assisted laparoscopic prostatectomy (RALP) has been demonstrated [5] to shorten this learning curve by leveraging the wristed instruments and the 3-D endoscopic camera of the telerobotic surgical system, usually the da Vinci surgical system, to achieve intuitive operation [8]. However, the endoscopic camera cannot localize the prostate lesions nor visualize the sub-surface anatomy of the prostate gland. Therefore, a complementary medical imaging modality is necessary to facilitate RALP.

Transrectal ultrasound (TRUS) facilitates the intraoperative identification of vital anatomic structures of the prostate gland and adjacent tissues, e.g., peripheral nerve and neurovascular bundles (NVBs), in a non-invasive and real-time manner [9]. TRUS provides precise ultrasound images of tissue pointed by the surgeon with the instrument. A robotic transrectal ultrasound system for prostatectomy was presented to alleviate the inconvenience of manipulating the TRUS and has been adopted in RALP studies [10–12]. The da Vinci and TRUS coordinate systems are registered by touching the tissue with the instrument tip at three or more different locations. However, the search process for each location is manual, time-consuming, and requires an additional skilled operator in the operating room [11].

A very accurate alternative registration method between the TRUS and the da Vinci system involves the use of photoacoustic markers (PMs), which are generated by the so-called photoacoustic effect, i.e., the formation of acoustic waves when a pulsed laser source is absorbed by optical materials. Robot assisted photoacoustic imaging using the da Vinci surgical system was investigated in [13–16]. In [17], authors demonstrated that photoacoustic markers generated by a pulsed laser could be reconstructed as images representing the local optical absorption of tissue constituents. Cheng et al. [18]'s research proved the feasibility and advantages of PM in surgical guidance. Instead of physically touching the tissue, a photoacoustic (PA) approach using focused laser illumination as the PM was introduced in [19], in which the registration between the endoscopic stereo camera frame and the TRUS frame was performed using Horn's method [33]. While accurate, this registration is still time-consuming (as much as 7 min), because the TRUS is manually rotated to align its imaging plane to the laser spots. Driven by this limitation, an automatic image-based search for the photoacoustic markers is proposed in this paper.

The directivity pattern of clinical transducer was investigated in [20]. The PA image intensity is proportional to the distance between the PM and the TRUS imaging plane. When the PM is coplanar with the imaging plane of the TRUS transducer, the detected PA image intensity of the PM is maximum [21]. Therefore, we have a PA image intensity function, which should be unimodal and symmetric. Based on this, the key component of our work is to conduct a one-dimensional search in the presence of PA imaging to find the position corresponding to the maximum PA intensity.

The numerical optimization of unimodal functions in one-dimensional (1-D) search space is a common topic in many engineering fields [22–24]. Introducing an efficient algorithm to replace the exhaustive search will significantly shorten the duration of the registration. The Golden Section Search (GSS) algorithm [25] is a classic approach to solve this maximization problem. Similarly, Ternary Search (TS) algorithm is a widely used divide-and-conquer algorithm for searching the maximum of a unimodal function [26,27].

In this paper, we propose a novel automatic search module to localize the PMs using the robotized transrectal ultrasound actuator in robot-assisted radical laparoscopic prostatectomy. We simulate and experimentally evaluate several optimization algorithms. The proposed framework outperforms the manual search in terms of time and required technical staff.

## 2. Method

### *2.1 Problem Definition*

In our RALP system depicted in Fig. 1. (a), a laser source generates a PM on the surface of the tissue. This PM can be detected by a TRUS beneath the tissue. This detected intensity of the PM achieves a maximum when the US imaging plane is coplanar with the PM, namely the TRUS is at the angle $\theta_p$, and attenuates when the imaging plane moving away from $\theta_p$. As described in Fig. 1. (b), there is a PA signal intensity function related to the rotation angle of the TRUS. Theoretically, this intensity function is unimodal and symmetric if noise is not considered [20], and we are searching for its peak. We can drive the motorized TRUS to rotate within -35$^o$ to 35$^o$, providing enough coverage for the prostate imaging, and synchronously calculate PA signal intensity at each location using the US+PA imaging system. In this way, we will investigate efficient algorithms to solve this problem compared to the manual scanning.

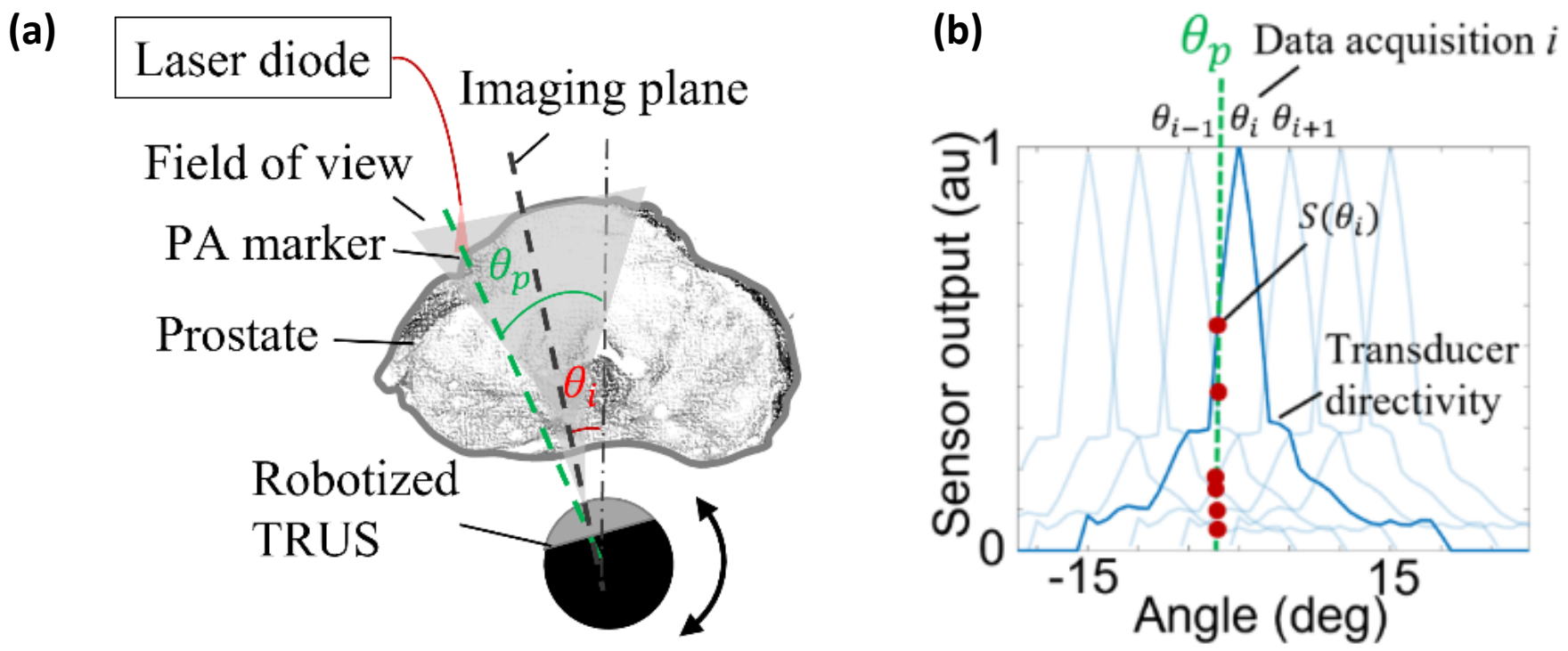

**Fig. 1.** (a) The laser illuminates a small spot (PM). The motorized TRUS is located at $\theta_i$. The goal is to find $\theta_p$ of the PA marker. (b) The TRUS also receives out-of-plane signals. Therefore, PA signals will be recorded at different angles.

### *2.2 Weighted Average*

The laser illuminates a small spot as shown in Fig. 1. (a). The motorized TRUS rotates from -35° to 35° in equal increments of $\theta_d$ at $\theta_i$s where $i = 1, ..., N$ for a full scan. The goal is to find $\theta_p$ in the TRUS coordinate system. The TRUS also receives out-of-plane signals. The maximum PA signal intensity, $S(\theta_i)$, is recorded for each $\theta_i$, described in Fig. 1. (b). As the directivity pattern of the TRUS transducer with respect to the incoming waves is symmetric, a simple weighted average can be used to localize the PA marker, $\theta_p$:

$$\theta_p = \sum \theta_i S(\theta_i) \,/ \sum S(\theta_i),\, i = 1,...,N \quad (1.)$$

Note that we need to set a threshold to filter out the samples with very low or no PA signal intensity. This helps to reduce noise and improve localization accuracy.

### *2.3 Golden Section Search and Ternary Search*

The Golden Section Search (GSS) is an efficient algorithm for finding the extremum (maximum or minimum) of a unimodal function. The GSS has been widely used in many applications because of its rapid convergence, without requiring that the objective function be continuous. In the GSS algorithm, we compare a pair of observation at two interior points $\theta_{m,1}$, $\theta_{m,2} \in R$ satisfying $\theta_l < \theta_{m,1} < \theta_{m,2} < \theta_r$ for each iteration. The two interior points partition the search interval in terms of the golden section ratio $r$. Here, we set the initial value of $\theta_l$ and $\theta_r$ as -35° and 35°. They are updated at each iteration and eventually convergence towards $\theta_p$. The interior points $\theta_{m,1}$ and $\theta_{m,2}$ are computed as follows:

$$r = \frac{\sqrt{5}-1}{2} \approx 0.618 \quad (2.)$$

$$\theta_{m,1} = r\theta_l + (1 - r)\theta_r \quad (3.)$$

$$\theta_{m,2} = (1 - r)\theta_l + r\theta_r \quad (4.)$$

The strategy of GSS can be stated as follows: if $S(\theta_{m,1}) \geq S(\theta_{m,2})$, the maxima must be in the interval $[\theta_l, \theta_{m,2}]$. Likewise, if $S(\theta_{m,1}) < S(\theta_{m,2})$, the maxima must be in the interval $[\theta_{m,1}, \theta_r]$. Subsequently, we compute two new interior points in the interval mentioned above for the next iteration. At every iteration, we calculate the interior points in this manner and keep narrowing the search interval until meeting the termination condition. Theoretically, the iteration would terminate when $\theta_l \geq \theta_r$ is met. In practice, however, we can set a termination condition to boost the speed of convergence. Once the search interval is less than this termination condition, we stop the iteration and take $\theta_p = (\theta_l + \theta_r)/2$ as the output location. If we set $r = 1/3$ instead of the golden section ratio of 0.618, the algorithm turns into the Ternary

Search (TS). Except for the scalar *r* used for interior points calculation, the other parts of the TS are the same as the GSS.

### 2.4 Simulation Validation

For the simulation, the experimental directivity pattern and PA model developed in [20] was used. The PA signals were generated by MATLAB k-Wave toolbox and White Gaussian noise was added to them. The intensity of PA signals was normalized. The reference angles of PA markers were located at 1000 random locations within -35° to 35°, and PA signal was recorded at the desired positions based on the proposed search algorithms. Here, we compared the performance of all three search algorithms with different numbers of data acquisitions. With the WA algorithm, laser spots were located at a random position and PA signals were detected at several $\theta_i$s spaced at certain $\theta_d$, determined by the number of acquisitions *N*:

$$\theta_d = 70^o/(N-1)\#(5.)$$

So, as the number of acquisitions increases, $\theta_d$ decreases. Note that the number of acquisitions cannot be too small because there may be no PA signals generated. In the GSS and TS algorithms, the angle of the TRUS is determined by the policy introduced in 2.3 and the iteration/acquisition can be continued until the desired accuracy is met. We tested the performance of these two algorithms when acquiring data 2-28 times.

### 2.5 Experiment Validation

In addition to the simulation validation, we conducted experimental studies to illustrate the advantages of our proposed method. In our experiment, a pulsed laser diode illuminated random locations on the tissue surface. To validate our approach, we conducted two types of analysis: offline and online. For the offline analysis, we conducted experiments to collect data by scanning the entire rotation interval with a very small step size (~0.5°), enabling us to have a bank of PA data for our investigation. We simulated the online search based on this collected data. The numerical precision of the motor rotation angle was made consistent with the online analysis by interpolating the original offline data. For the online analysis, the algorithms were investigated in a real-time implementation where each of the search algorithms were integrated into the robotic and imaging systems. As a result, PA data were acquired at the locations that the algorithm asked for, and the output was compared with the reference acquired from a manual search.

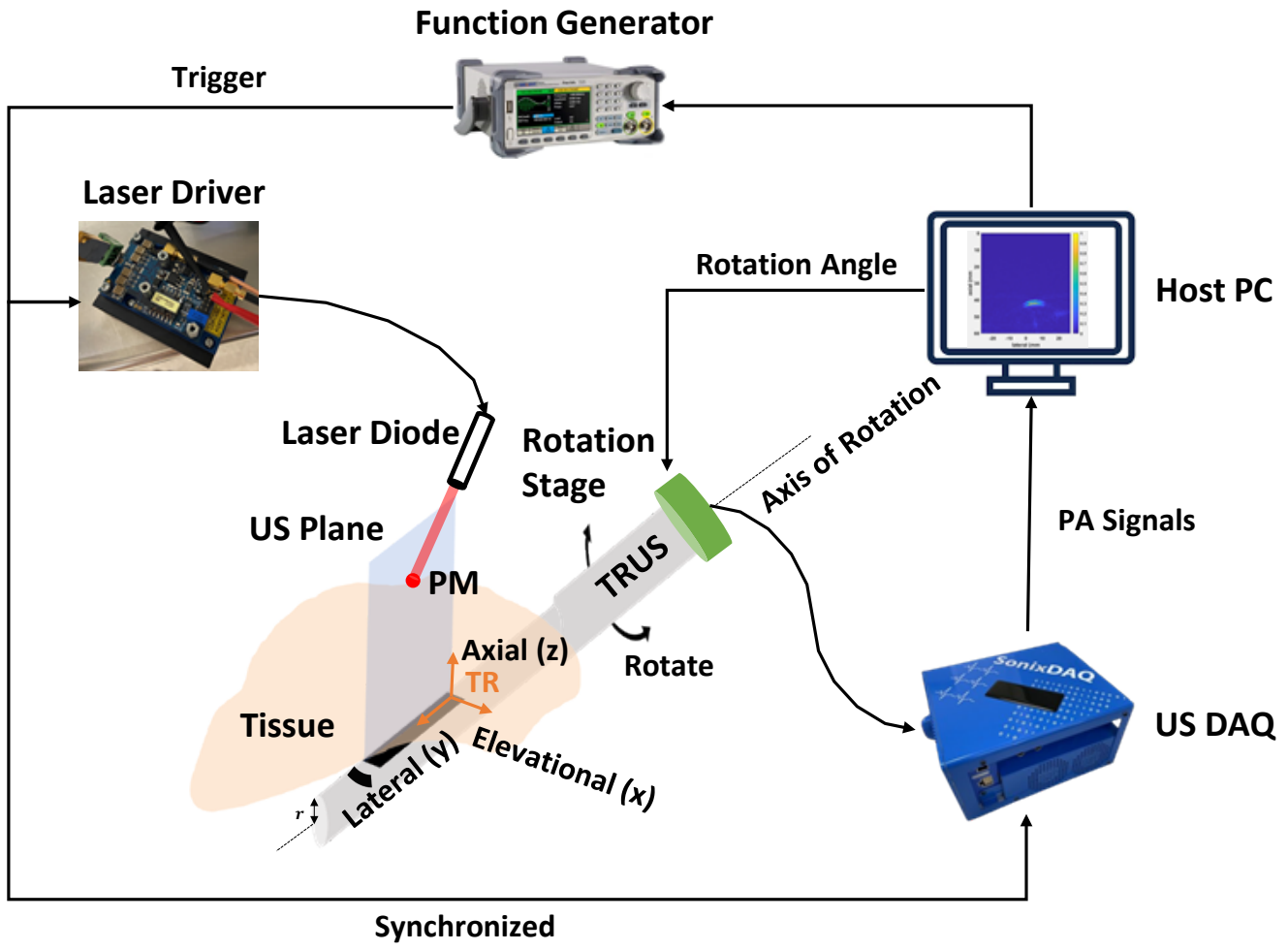

**Fig. 2.** Schematic diagram of the overall system composition. It also describes the TRUS coordinate system (TR). The *y*-axis is parallel with the axis of rotation of the TRUS.

*A. System Architecture*

The overall system depicted in Fig. 2. comprises of several modules including the main host computer, pulsed laser system, TRUS actuator control module, and PA+US imaging system. The imaging system was described in [19]. Here, we emphasize the modules related to the automatic implementation of the search algorithms. In the pulse laser system, the 785 nm pulse laser diode (QSP-785-4, QPhotonics LLC, Ann Arbor, MI, USA) is excited by a laser driver (LDP-V, PicoLAS GmbH, Germany) to generate a 2µJ/pulse laser beam. The PA+US imaging module synchronously acquires PA signals. The generated PA signals can be detected by a TRUS transducer (BPL 9-5/55, BK Medical, Peabody, MA, USA) and a SonixDAQ (Ultrasonix Medical Corp., Canada). The actuator control module can drive the TRUS transducer to rotate through an actuator (rotation stage, PRM1Z8, Thorlabs, Newton, NJ, USA) rigidly attached to it. Real-time communication between each component in the system is based on the Robot Operating System (ROS).

*B. Experimental Setup*

The experimental set-up is shown in Fig. 3(a). An *ex vivo* bovine liver tissue was supported by two pieces of ultrasound gel pad (Parker Laboratories, Fairfield, NJ, USA) with a total thickness of 5 cm to mimic the prostate gland. A laser diode held by a clip was placed close to the surface of the *ex vivo* tissue. The laser's location and its surrounding region (5mm × 5mm area) were stained with India ink (Dr. Ph. Martin, Oceanside, CA, USA) to increase laser absorption. A TRUS transducer motorized by an actuator was placed beneath the *ex vivo* tissue and ultrasound pads to detect the PM. Here, we tested the proposed approach on a homogenous black plastisol phantom, and three types of *ex vivo* tissue, bovine liver, muscle, and chicken breast, to demonstrate its feasibility. In one case, in accordance with the prostate nerve sensing [28], the *ex vivo* bovine muscle was stained by indocyanine green (ICG) dye (TCI, Tokyo, Japan). As illustrated in Fig. 3. (b), the PA signals arranged on a 3 × 3 grid were collected on each type of material.

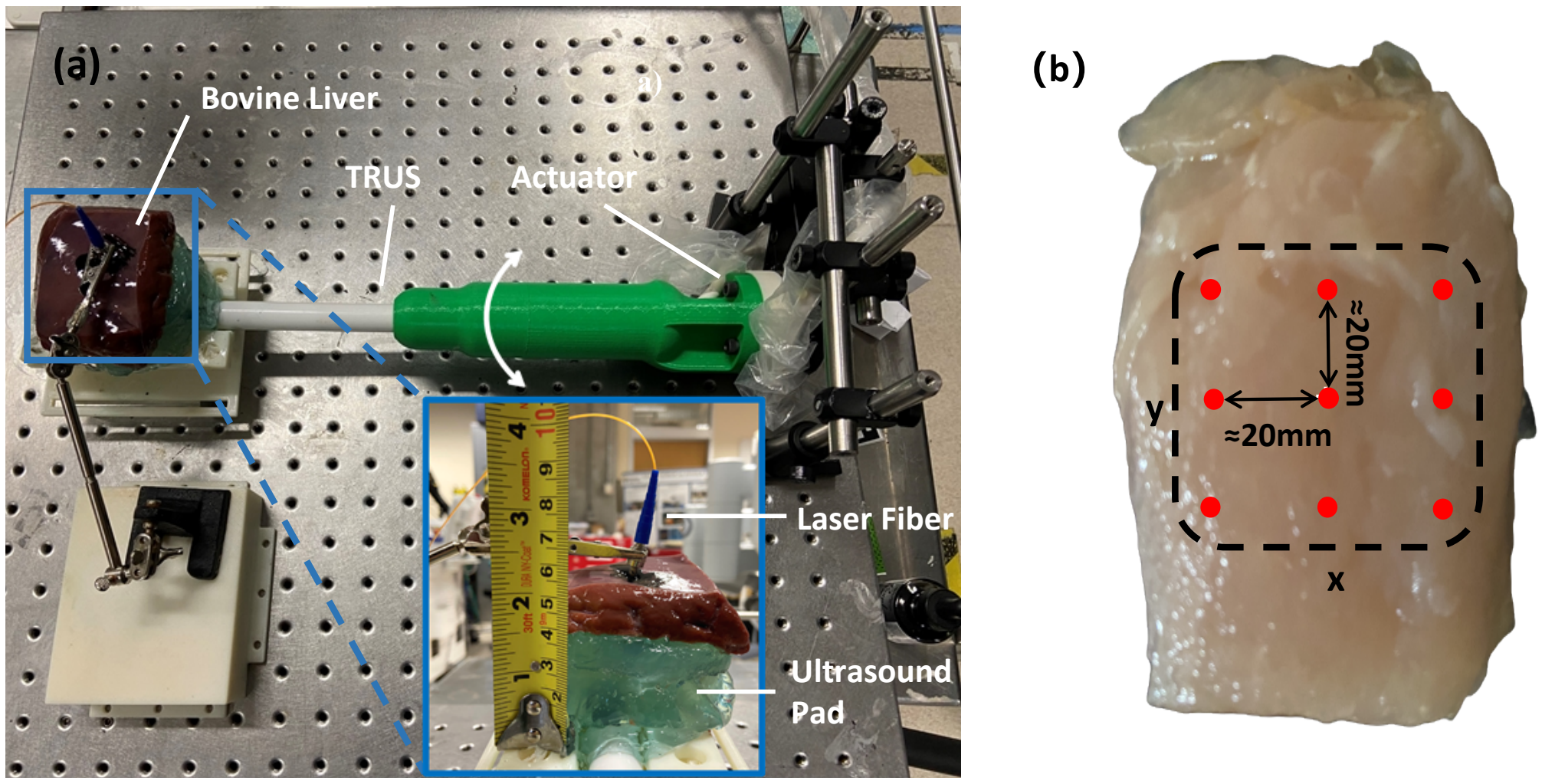

**Fig. 3.** (a) The offline analysis experimental set-up with bovine liver *ex vivo* tissue stained with India ink. (b) The conceptual illustration of the laser spot location on *ex vivo* chicken breast tissue from the top view of the *x*-*y* plane. The red dots indicate the laser spot, and a 3×3 grid roughly spaced at 20mm in the *x*-axis and *y*-axis directions, respectively, of the *x*-*y* plane of the TRUS coordinate system. For other *ex vivo* tissues and plastisol phantom, the laser spot location is arranged in the same way.

*C. Online Search implementation*

We integrated the automatic search module into the overall TRUS-da Vinci robotic system using ROS and used a LabVIEW script [19] to control the rotation stage. To integrate this TRUS controller into our ROS communication system, we used the ROS for LabVIEW toolkit software [29] developed by the Tufts University Center for Engineering Education and Outreach. We created a node for the TRUS controller, as well as a publisher and a subscriber to receive the angle that the TRUS should rotate to and return the actual rotated angle. For online analysis, there was no need to exhaustively scan the entire rotation range with a small step size to collect data as in our offline analysis. The TRUS only rotated to specified locations that the algorithm asked for and acquired PA signals from the data acquisition system at these locations. Here, we tested the proposed online automatic search module on a black plastisol phantom. Like in the offline analysis case, we tested each search algorithm on the $3 \times 3$ grid of locations and the results were compared with manually found reference locations.

*D. Tool Tip Detection - Sensitive Range*

When using the da Vinci instrument tool tip for registration, a key stage in this procedure is the da Vinci surgical instrument tip detection [30]. Similar to the PM intensity, the tip intensity is the highest when the US imaging plane is coplanar with the surgical instrument tip. It decreases and eventually disappears as the US imaging plane deviates from the coplanar position. Consequently, we only need to make sure the surgical instrument tip is within the range (termed sensitive range) that can be detected by the US, not necessarily to be exactly coplanar with the US imaging plane.

We experimentally investigated the sensitive range when detecting the da Vinci surgical instrument tip. As shown in Fig. 4., the da Vinci instrument tip held on a robotic arm was pressed against the surface of a piece of *ex vivo* chicken breast. We rotated the TRUS clockwise with a step size of $1°$ starting with the reference location and recorded the corresponding B-mode image at each location. The range from the reference location to the location where the tip disappeared (shown in Fig. 9. (b) and (c)) was half of the sensitive range. Assuming that the function of detected tip intensity versus TRUS rotation location is symmetric, the other half sensitive range was inferred by flipping along the reference location. Here, we roughly calculated the entire sensitive range by adding these two half sensitive ranges. Three locations of the da Vinci surgical instrument tip spaced at about 2cm in $x$-axis were tested in this experiment.

For the B-mode image processing, we first segmented the da Vinci instrument tip by leveraging GrabCut [31], following to find the centroid of this connected instrument tip area, and finally taking the pixel value of this centroid as the instrument tip intensity. Let $\theta_{high}$ denote the location with the highest tip intensity, and $\theta_{low}$ denote the location with the lowest tip intensity that can be detected. Thus, half of the sensitive range can be defined as the absolute value of these two locations, namely $|\theta_{high} - \theta_{low}|$.

## 3. Result

The proposed automatic search module was validated by both simulation and experiment (including offline analysis and online analysis). The performance of various search algorithms was evaluated in terms of localization accuracy and time efficiency. For the evaluation of localization accuracy, we measured the localization error which was defined as the absolute value of the difference between the reference location and optimal output location. To assess time efficiency, we adopt the number of data acquisitions as the metric due to most of the time in the proposed search procedure being used to acquire data. In the following sections, we will illustrate and discuss the results of each in detail.

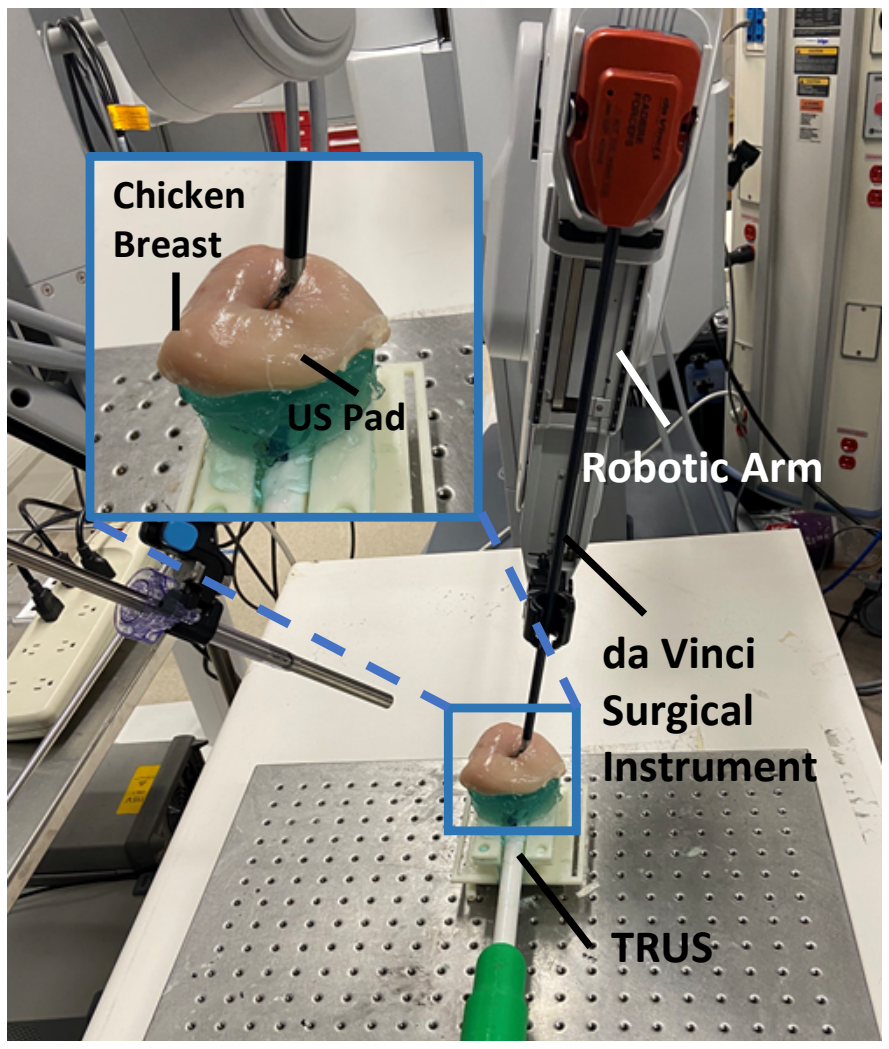

**Fig. 4.** The experimental setup for investigating the sensitive range of da Vinci surgical instrument tip detection.

### *3.1 Simulation Results*

For the TS and GSS algorithm, we tested a set of number of acquisitions from 2 to 28 in increments of 2. Likewise, we tested a set of number of acquisitions from 8 to 28 for the WA algorithm to ensure that at least one valid PA signal can be acquired at the proposed equal intervals. We set a fixed threshold of 0.2 for all experiments in the WA algorithm. The average error after 1000 trials changed with the number of acquisitions as shown in Fig. 5. The error of TS and GSS algorithms rapidly dropped at the beginning and converged to a small value, while the error of WA did not change much with the number of acquisitions. When conducting 28 acquisitions, the average errors of the TS and GSS algorithm were 0.06° and 0.01°, respectively. In the WA algorithm, the average error when acquiring different numbers of PA signals was $0.47° \pm 0.11°$.

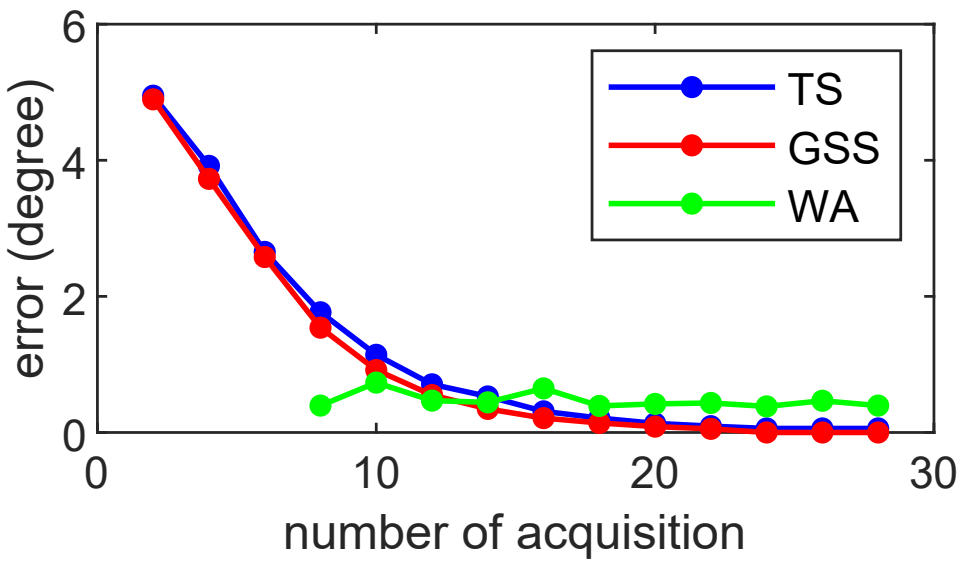

**Fig. 5.** The average error of 1000 experiments using various algorithms in the simulation study. Red dots, blue dots, and green dots represent the error of TS, GSS, and WA algorithm.

### *3.2 Offline Analysis Results*

In our offline analysis, we interpolated each set of data to achieve the same resolution as the simulated data. Fig. 6. depicts the average error versus the number of acquisitions on various materials resulting from the offline analysis. Table 1 summarizes the error of each algorithm when acquiring the PA signal 16 times. Note that we only obtained one set of data on bovine muscle stained by ICG, so the results of this test did not have a standard deviation.

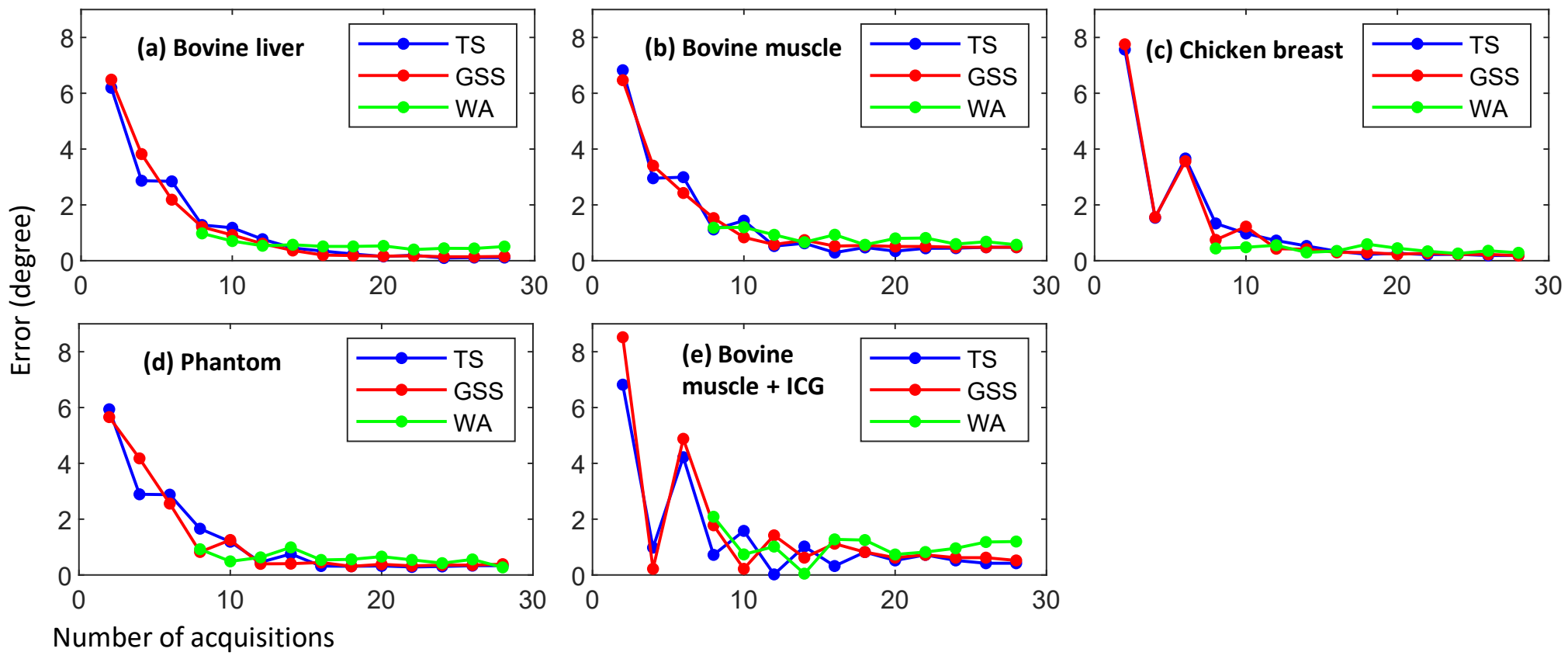

**Fig. 6.** The average error of offline analysis versus the number of acquisitions. The sub-figures (a) - (e) indicate the offline analysis results on bovine liver, muscle, chicken breast, phantom and bovine muscle stained with ICG, respectively.

**Table 1. Average Error of Various Algorithms in Offline Analysis Terminating After 16 Times of Acquisitions**

| | Average Error (Degree) | | | | |
|---|---|---|---|---|---|
| Algorithms | Bovine Liver | Bovine Muscle | Chicken Breast | Plastisol Phantom | Bovine Muscle +ICG |
| WA | $0.51 \pm 0.36$ | $0.93 \pm 0.83$ | $0.35 \pm 0.23$ | $0.54 \pm 0.41$ | 1.28 |
| TS | $0.34 \pm 0.27$ | $\mathbf{0.29 \pm 0.22}$ | $0.34 \pm 0.20$ | $\mathbf{0.32 \pm 0.22}$ | **0.32** |
| GSS | $\mathbf{0.20 \pm 0.19}$ | $0.52 \pm 0.61$ | $\mathbf{0.30 \pm 0.24}$ | $0.45 \pm 0.40$ | 1.12 |

### *3.3 Online Analysis Results*

In online analysis, each algorithm described earlier was integrated into the TRUS-da Vinci robotic system. When conducting the online analysis, the TRUS automatically rotated to the position specified by the algorithm, and finally stopped at $\theta_p$ calculated by this algorithm. Fig. 7. illustrates the comparison of the average errors of TS and GSS algorithms with varying numbers of acquisitions. The average errors of these two algorithms decrease with an increase in the number of acquisitions. For the TS and GSS algorithms, the specific location of the TRUS at each number of acquisitions is illustrated in Fig. 8. The GSS and TS algorithm achieved minimum error, $0.48^{o} \pm 0.30°$ and $0.29^{o} \pm 0.30°$, when acquiring 28 times data. When adopting WA, we took 5° as the step size, that is, acquiring 15 times PA signals to cover the entire space. However, in the online analysis, we can terminate the search as long as the entire peak of the intensity is recorded and passed. We can often complete the WA search by collecting data fewer than 15 times. The average error of WA was $0.53^{o} \pm 0.30^{o}$ with 8.7 data acquisitions. Table 2 is the comparison of the average error of various algorithms when taking a similar number of acquisitions.

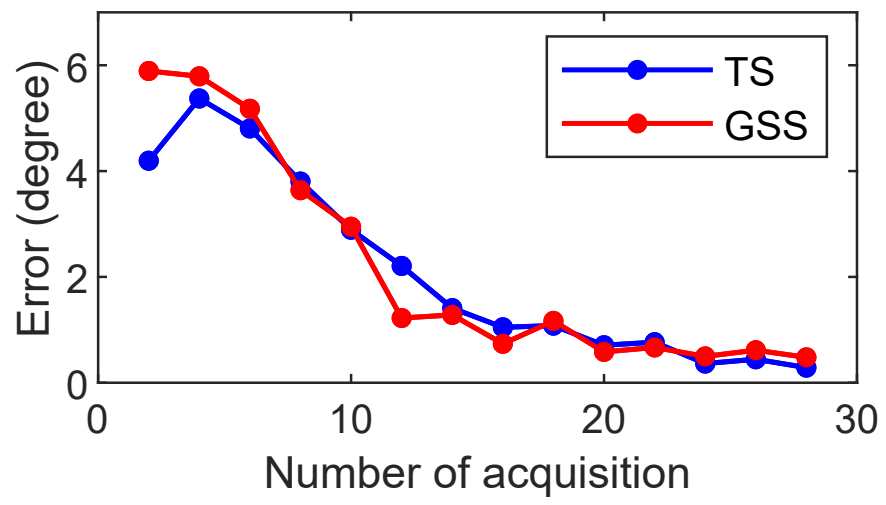

**Fig. 7.** Average error of TS and GSS with a varying number of acquisitions in online analysis. The blue bar and orange bar indicate the TS and GSS algorithms respectively.

**Table 2. Average Error of Various Algorithms in Online Analysis**

| Algorithms | Average Error | Number of Acquisitions |
|---|---|---|
| WA | 0.53 ± 0.30 | 8.7 |
| TS | 1.05 ± 1.04 | 16 |
| GSS | 0.74 ± 0.58 | 16 |

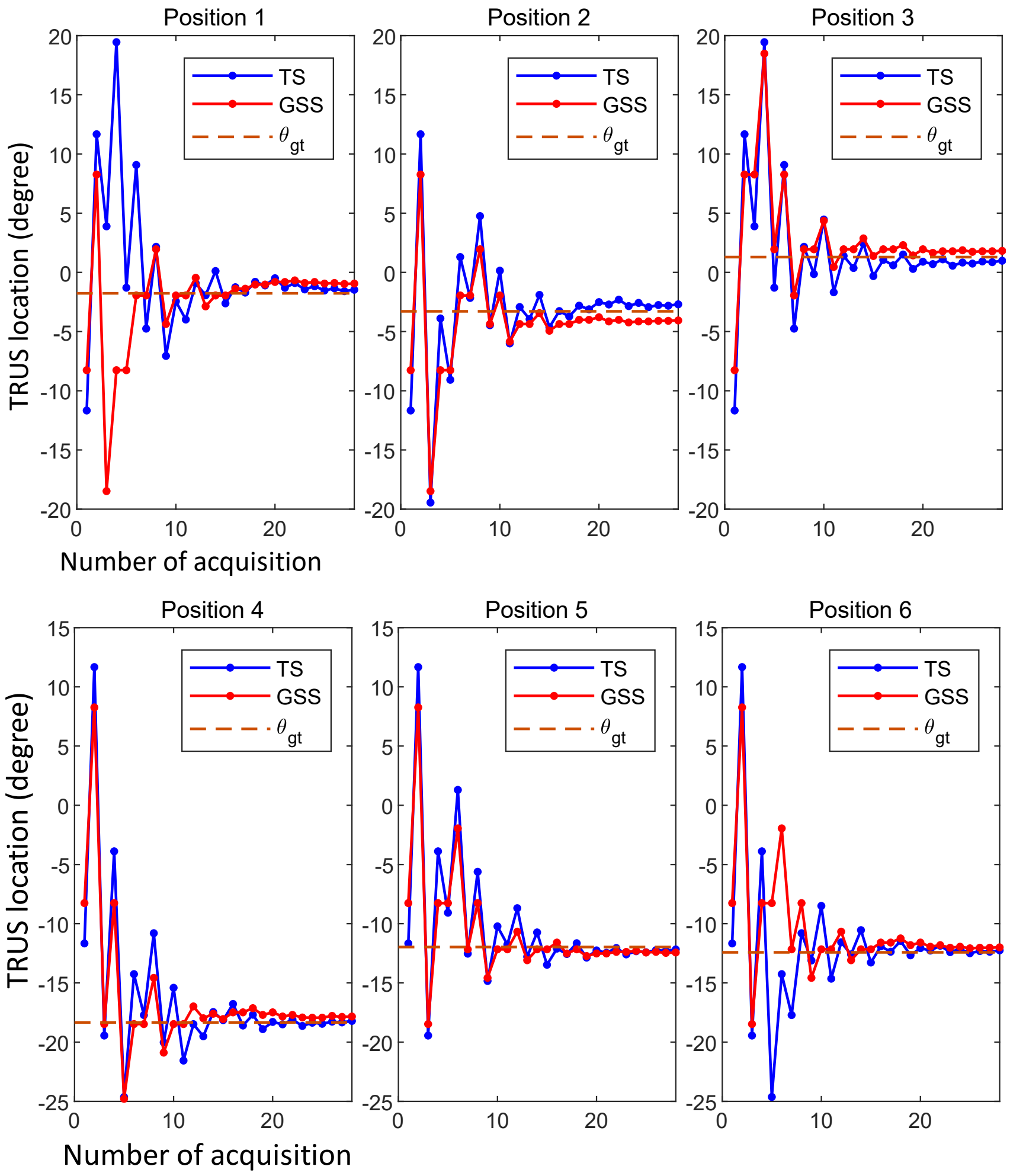

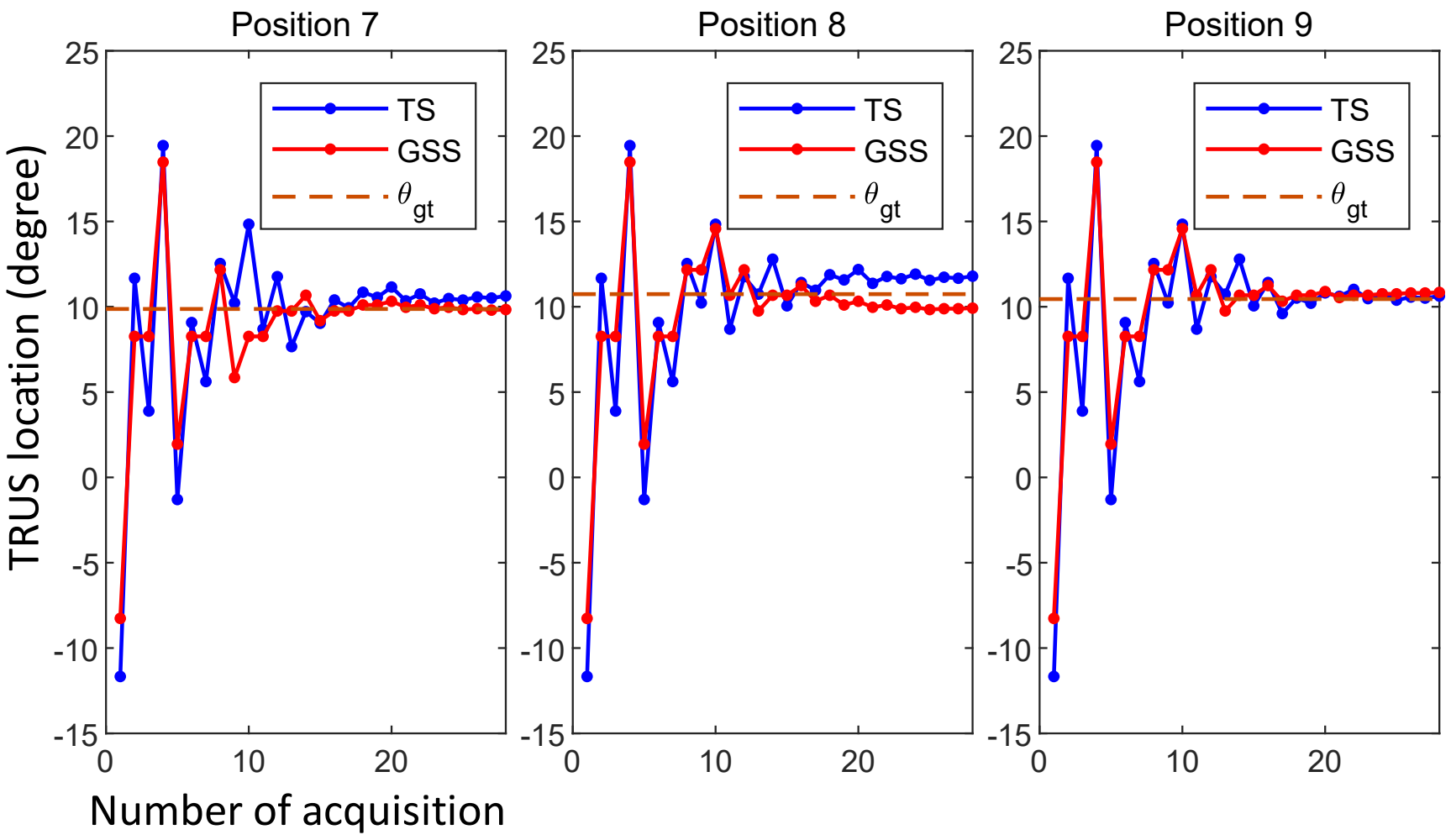

**Fig. 8.** TRUS location during the TS and GSS algorithms. The blue dots, red dots, and orange dashed line refer to the locations in TS and GSS as well as the reference location $\theta_{gt}$. The position 1 to 9 were located on a 3 × 3 grid as described in Fig. 3. (b).

### *3.4 Sensitive Range Results*

We pressed the da Vinci surgical instrument tip against the *ex vivo* chicken breast tissue surface at three locations. For each tip location, a set of B-mode images were recorded with the TRUS rotated in steps of 1° (Fig. 9. (a) and (d) show one of the B-mode images sets and the corresponding tip intensity curve). We took the position corresponding to the strongest and weakest tip intensity that can be detected as half of the sensitive range. In our experiment, half of the sensitive range of the three locations were 5°, 4°, and 4° respectively, and the average value was 4.33° ± 0.58°.

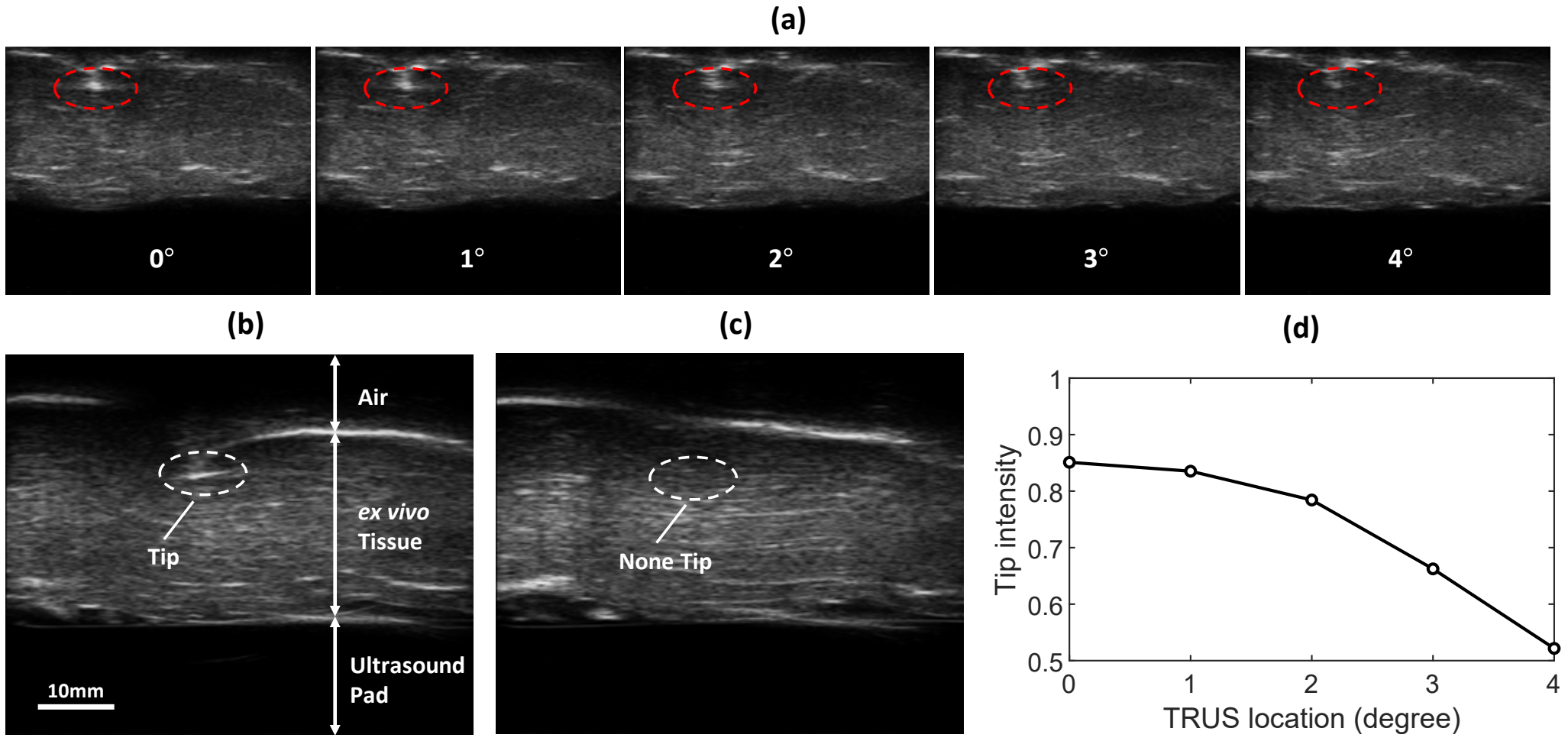

**Fig. 9.** (a) A set of B-mode images with varying tip intensity. The da Vinci surgical instrument tip is within the red dashed circle. The reference location is 0°. (b) The B-mode image at the reference location. It can be partitioned into 3 layers: from top to bottom are air, *ex vivo* tissue and ultrasound pad. (c) The B-mode image at a location transcending the sensitive range. (d) The tip intensity versus TRUS location. As the TRUS location moves away from the reference location 0°, the corresponding tip intensity keeps reducing.

## 4. Discussion

In terms of the experimental results described in 3.4, the sensitive range for the da Vinci surgical instrument tip detection is about 8.66°. In fact, we can obtain a much lower error ($0.53° \pm 0.30°$) using the WA algorithm in our study just acquiring PA signals eight times. Consequently, the accuracy of all three algorithms should be considered precise and sufficient compared to the sensitive range of tip detection. Regarding the three algorithms adopted in this paper, TS and GSS output the most precise PM location on *ex vivo* tissues when acquiring data 28 times.

The proposed search algorithms have the potential to be used as a part of the control platform for the surgical tool tracking and TRUS servoing tasks. In online analysis, and with only acquiring two PA signals, the average error was 4.20° and 5.89° when adopting the TS and GSS algorithm starting at the far left (-35°) and right (+35°) of the boundary which is reasonable when compared to the required 8.66°. However, this accuracy will significantly improve with more data acquisitions as illustrated in this paper. Besides, the TS and GSS tend to converge toward and settle at the target location $\theta_p$. This property makes these two algorithms potentially applicable in tool tracking procedures and will be further investigated in our future studies.

From the perspective of time, the primary limitation of the proposed technique is the relatively long time of acquiring PA signals from SonixDAQ. In this work, we utilized the API of SonixDAQ to read raw PA signals from the device and read these raw data files to do beamforming and reconstruction. This manner typically takes about 10 seconds. Driven by this limitation, our group is developing a real-time PA imaging based on SPARE (Synthetic-aperture Based Photoacoustic Re-beamforming [32]). This can essentially benefit the time efficiency of the PM-based RAPL system. Future work will integrate SPARE into the system to ameliorate the time-consuming PA signals acquisition and improve the system's efficiency. In that case, the only time we need to consider is the time of TRUS rotation. When acquiring data for 16 times, the total rotation travel in TS and GSS are about 36° and 24°, respectively. Thus, the estimated time of rotation is within 1-2 seconds, which is much less than the 2-3 minutes of manual search reported in [30] and close to a real-time search approach.

## 5. Conclusion

In this study, we presented an automatic search method for PM to deal with the drawbacks of manual search procedures. We cast the PM search procedure as an optimization problem of a 1-D unimodal function and investigated three algorithms to address this problem. Simulation studies and experiments on various *ex vivo* tissues and phantom validated the validity of our proposed method. In simulation studies, all three algorithms perform well. In offline analysis, the average error of the TS and GSS algorithm on all materials was below 0.5° when acquiring 28 PA acquisitions. In online analysis, the average error of the TS and GSS algorithm achieved $0.29^{o} \pm 0.31^{o}$ and $0.48^{o} \pm 0.32^{o}$ after 28 acquisitions, while the WA achieved $0.53^{o} \pm 0.30^{o}$ using only 8.7 times acquisitions. In summary, the proposed technique significantly improved the system's degree of automation and alleviated the labor required, while achieving good search accuracy. In future work, the proposed search method will be combined with real-time photoacoustic imaging to significantly improve the time efficiency of the PM-based RAPL.

**Funding.** This work was supported in part by NSF Career Award 1653322, NIH R01-CA134675, the Johns Hopkins University internal funds, Canadian Institutes of Health Research (CIHR), CA Laszlo Chair in Biomedical Engineering held by Professor Salcudean, and an equipment loan from Intuitive Surgical, Inc.

**Acknowledgments.** We would like to acknowledge to our sponsors and funding agencies: Funding from the National Science Foundation Career Award 1653322, National Institute of Health R01-CA134675, the Johns Hopkins University internal funds, Canadian Institutes of Health Research, CA Laszlo Chair in Biomedical Engineering held by Professor Salcudean, and Intuitive Surgical for the equipment support.

**Disclosures.** The authors declare no conflicts of interest.

**Data availability.** Data underlying the results presented in this paper are not publicly available at this time but may be obtained from the authors upon reasonable request.